\documentclass{article}

\PassOptionsToPackage{numbers, compress}{natbib}

\usepackage[final]{neurips_2024}

\usepackage[utf8]{inputenc} 
\usepackage[T1]{fontenc}    
\usepackage{hyperref}       
\usepackage{url}            
\usepackage{booktabs}       
\usepackage{amsfonts}       
\usepackage{nicefrac}       
\usepackage{microtype}      
\usepackage{xcolor}         
\usepackage{amsmath}
\usepackage{graphicx}
\usepackage{multirow}
\usepackage{booktabs}
\usepackage{siunitx} 

\title{Towards Privacy-Preserving Medical Imaging: Federated Learning with Differential Privacy and Secure Aggregation Using a Modified ResNet Architecture}

\author{%
  Mohamad Haj Fares \\
  Department of Computer Engineering \\
  Istanbul University-Cerrahpasa \\
  Istanbul, Türkiye \\
  \texttt{ai.mohres@gmail.com} \\
  \And
  Ahmed Mohamed Saad Emam Saad \\
  School of Computing \\
  Queen's University \\
  Kingston, ON K7L 3N6, Canada \\
  \texttt{a.saad@queensu.ca} \\
}

\begin{document}

\maketitle

\begin{abstract}
  With increasing concerns over privacy in healthcare, especially for sensitive medical data, this research introduces a federated learning framework that combines local differential privacy and secure aggregation using Secure Multi-Party Computation for medical image classification. Further, we propose DPResNet, a modified ResNet architecture optimized for differential privacy. Leveraging the BloodMNIST benchmark dataset, we simulate a realistic data-sharing environment across different hospitals, addressing the distinct privacy challenges posed by federated healthcare data. Experimental results indicate that our privacy-preserving federated model achieves accuracy levels close to non-private models, surpassing traditional approaches while maintaining strict data confidentiality. By enhancing the privacy, efficiency, and reliability of healthcare data management, our approach offers substantial benefits to patients, healthcare providers, and the broader healthcare ecosystem.
\end{abstract}

\section{Introduction}

Deep learning for medical imaging faces significant challenges in leveraging advancements from other fields of computer vision due to privacy concerns and limited data access \cite{adnan2022federated, sheller2020federated}. Privacy-enhancing technologies, such as Federated Learning (FL) and Differential Privacy (DP), provide viable solutions by enabling insights from sensitive data while safeguarding individual privacy \cite{adnan2022federated, mcmahan2018general}. FL enables decentralized model training directly on user devices or within hospital systems, removing the need for central data aggregation and thereby reducing the risk of data breaches \cite{lim2020federated, lalitha2018fully}. In this setup, data remains stored locally, while a central server coordinates training by exchanging model updates, ensuring data confidentiality throughout the process.

DP further strengthens privacy by introducing controlled noise to data, which masks specific information while preserving model accuracy \cite{dwork2008differential, choquette2024amplified, choquette2022multi, kairouz2021practical}. This method includes gradient clipping to limit individual data contributions \cite{andrew2021differentially}, followed by the addition of noise proportional to data sensitivity, making it challenging for adversaries to reconstruct individual data points, even if they gain access to the trained model. To enhance security in FL further, Secure Multi-Party Computation (SMPC) \cite{yao1982protocols, mohassel2017secureml} is employed for Secure Aggregation (SecAgg), ensuring that individual model updates remain private from both the central server and any unauthorized parties during transmission.

In this paper, we introduce a privacy-preserving FL framework that combines the Federated Averaging (FedAvg) algorithm \cite{mcmahan2017communication} with local DP, using fixed gradient clipping, alongside SMPC-based SecAgg. Additionally, we propose DPResNet, a modified ResNet-9 architecture optimized for differential privacy within federated settings, further enhancing model performance under privacy constraints. Applied to the BloodMNIST datasets, this approach simulates realistic data-sharing environments in medical imaging. Our results show that this framework achieves high accuracy while adhering to strict privacy standards, enabling secure and effective deployment of Deep Neural Networks (DNNs) in healthcare applications.

\paragraph{Contributions} Our main contributions are as follows:

\begin{itemize}
    \item We develop a privacy-preserving FL framework for medical imaging, integrating the FedAvg algorithm with local DP, including gradient clipping, and SMPC-based SecAgg to protect data confidentiality during model training and aggregation.

    \item We propose DPResNet, a modified ResNet architecture tailored for differential privacy in federated settings to enhance compatibility with privacy-preserving protocols.
    
    \item We apply this framework to the BloodMNIST dataset, creating a realistic simulation of multi-institutional data-sharing scenarios across decentralized healthcare environments.
    
    \item We evaluate the performance of our framework, demonstrating that it achieves competitive accuracy levels while maintaining strong privacy guarantees, close to those of non-private models.
\end{itemize}

\section{Related Work}
DP and FL are foundational for privacy-preserving medical imaging. DP-Stochastic Gradient Descent (DP-SGD) \cite{abadi_2016} introduced gradient clipping and noise addition to protect data privacy, with subsequent adaptations improving scalability and refining privacy-accuracy trade-offs \cite{yu2021large, papernot2018scalable}. Combined with FL, DP enables near-centralized performance without raw data aggregation in multi-institutional setups \cite{sheller2020federated, kaissis2021end}. However, model update transmission remains vulnerable to data leakage, which SMPC \cite{yao1982protocols, mohassel2017secureml} addresses by preserving privacy during aggregation, even when DP alone may fall short \cite{10.1007/978-3-030-32692-0_16, mcmahan2017communication}.

Personalized FL frameworks tackle non-IID data heterogeneity by customizing model weights per client \cite{zhang2020arxiv, luo2021arxiv}. Approaches like FedProx \cite{li2018arxiv} use regularization to handle data variability, informing cross-device FL approaches \cite{tan2022towards}. Domain-specific adaptations of DP for medical fields, including histopathology and brain imaging, often employ Gaussian noise but lack sensitivity clipping, affecting privacy guarantees \cite{lu2022MedicalImageAnalysis, silva2020fed}. Methods such as transformer-based pre-training and domain adaptation improve FL performance with non-IID data distributions \cite{Yan_2023, 10.1007/978-3-030-32692-0_16}.

FL optimization for resource-constrained edge devices involves methods like FedSup \cite{zhao2023fedsup}, incorporating neural architecture search and Bayesian Convolutional Neural Networks (BCNN) for uncertainty, and asynchronous aggregation to reduce communication costs \cite{kim2022arxiv, lim2020federated}. 

Despite these advancements, privacy vulnerabilities remain, with adversarial attacks, as demonstrated by MediLeak \cite{shi2024harvestingprivatemedicalimages}, and Generative Adversarial Networks (GANs) capable of reconstructing private data, underscoring the need for robust defenses \cite{hitaj2017deep, nasr2019comprehensive}. Although promising, FL with DP and SMPC faces challenges like privacy-accuracy trade-offs and a lack of standardized datasets, complicating reproducibility.

While FL with DP and SMPC-based aggregation shows promise, challenges like the privacy-accuracy trade-off and the scarcity of standardized datasets hinder reproducibility and benchmarking. Our study addresses these gaps by applying FL with DP and SMPC to the MedMNIST dataset, exploring privacy and accuracy trade-offs in privacy-preserving medical imaging.

\section{Methodology}
\subsection{Secure Federated Learning Framework}
Our framework combines FL with DP and Secure SMPC for privacy-preserving medical image classification. The FL objective is to train a model \(f_{\theta} : \mathcal{X} \rightarrow \mathcal{Y}\) with parameters \(\theta\) using distributed datasets \(\mathcal{D} = \{\mathcal{D}_i\}_{i=1}^N\) across \(N\) hospitals or devices, where data \(\mathcal{D}_i\) remains local. The federated optimization problem is defined as:

\begin{equation}
\min_{\theta} \sum_{i=1}^N p_i \cdot \mathcal{L}(\mathcal{D}_i; \theta),
\label{eq:fl_optimization}
\end{equation}

where \(p_i\) is the weight of client \(i\) and \(\mathcal{L}(\mathcal{D}_i; \theta)\) is the loss on the local data \(\mathcal{D}_i\). Each client trains locally and transmits updates \(\Delta \theta_i\) to the server for SecAgg.

\paragraph{Privacy Mechanisms}
To ensure privacy, clients apply gradient clipping to limit the sensitivity of updates. The clipped gradient is:

\begin{equation}
g_i^{\text{clipped}} = \frac{g_i}{\max\left(1, \frac{\|g_i\|}{C}\right)},
\label{eq:gradient_clipping}
\end{equation}

where \(C\) is the clipping norm. After clipping, Gaussian noise calibrated to an \((\epsilon, \delta)\)-DP guarantee is added:

\begin{equation}
\Delta \theta_i^{\text{DP}} = g_i^{\text{clipped}} + \mathcal{N}(0, \sigma^2 I),
\label{eq:noise_addition}
\end{equation}

where \(\sigma\) is the noise scale. These mechanisms ensure that individual contributions remain obfuscated in the transmitted updates.

\paragraph{Secure Aggregation}
To protect model updates during aggregation, SMPC aggregates the updates securely without exposing individual contributions. The aggregated model update is:

\begin{equation}
\theta_{t+1} = \text{SecAgg}\left(\{\Delta \theta_i^{\text{DP}}\}_{i=1}^N\right) = \sum_{i=1}^N p_i \cdot \Delta \theta_i^{\text{DP}},
\label{eq:secure_aggregation}
\end{equation}

where \(\text{SecAgg}(\cdot)\) denotes the Secure Aggregation operation. This protocol prevents an honest-but-curious server from accessing individual updates while enabling global model updates. 

\paragraph{Modified ResNet Architecture}
Our model, DPResNet, adapts ResNet-9 by replacing BatchNormalization with GroupNormalization (32 groups per layer) and removing max-pooling layers. GroupNormalization computes statistics over grouped channels, making it compatible with DP requirements while maintaining simplicity and effectiveness.

\paragraph{Overall Objective}
The overall objective integrates FL, DP, SMPC-based SecAgg, and the DPResNet architecture. This can be expressed as:

\begin{equation}
\min_{\theta^{\prime}} \sum_{i=1}^N p_i \cdot \mathbb{E} \left[ \mathcal{L}(\mathcal{D}_i; \theta^{\prime}) \right] \quad \text{subject to } (\epsilon, \delta)\text{-DP constraints}.
\label{eq:final_objective}
\end{equation}

Equations \eqref{eq:fl_optimization} through \eqref{eq:final_objective} encapsulate the integration of the proposed secure FL framework.

\subsection{Workflow and Experimental Setup}

\begin{figure}
  \centering
  \includegraphics[width=2\textwidth,height=0.22\textheight,keepaspectratio]{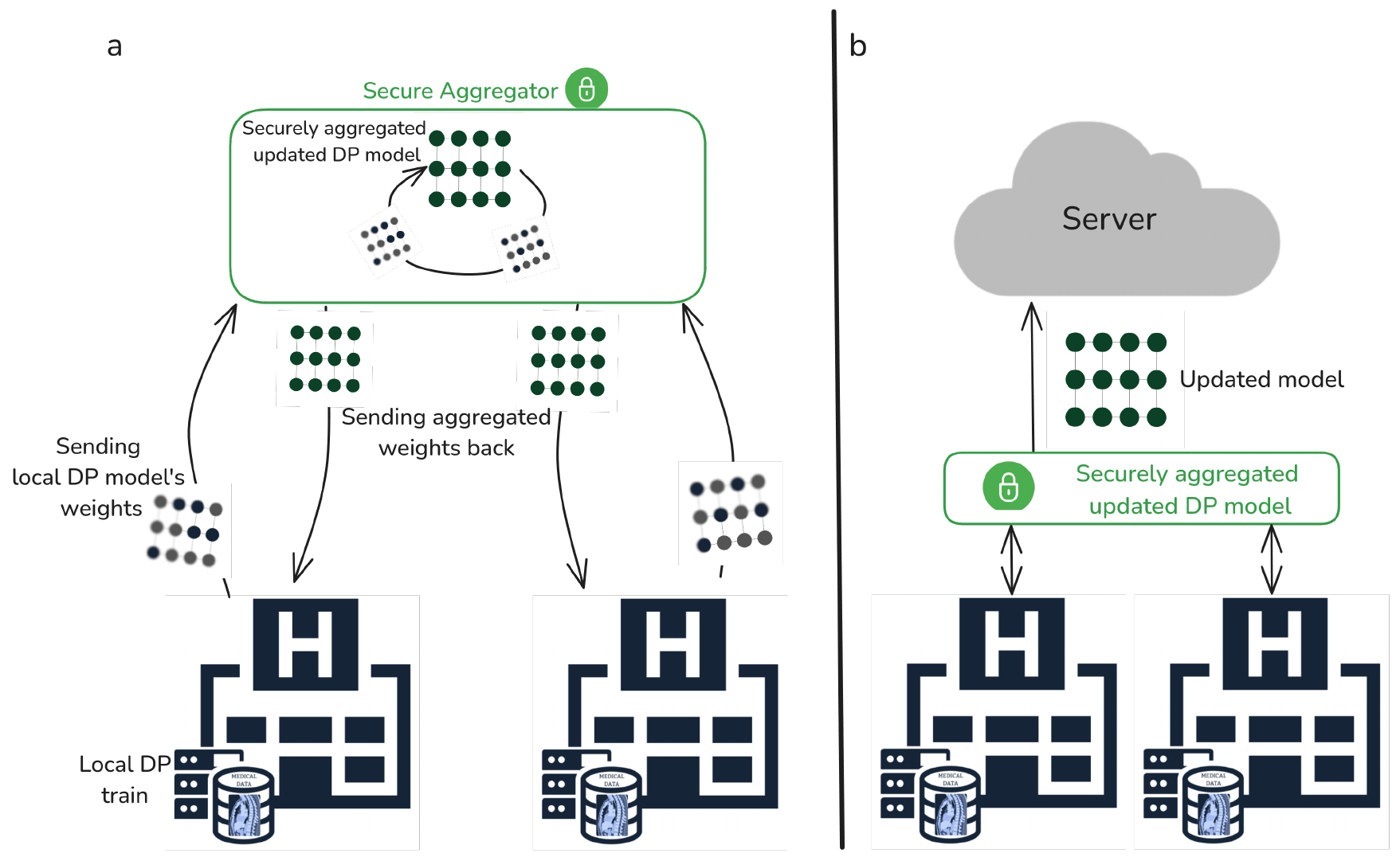}
  \caption{Illustration of the FL workflow. (a) Models are trained locally with differential privacy and aggregated securely using SecAgg+. (b) The final aggregated model (green) is updated after multiple rounds.}
  \label{fig:DPFLSA}
\end{figure}

To illustrate the workflow of our privacy-preserving framework, Figure~\ref{fig:DPFLSA} presents the interaction between hospitals during the training process. Each hospital \(i\) trains a local model \(f_{\theta_i}\) using a differentially private approach, ensuring that sensitive data \(\mathcal{D}_i\) remains protected.

Locally trained updates \(\Delta \theta_i^{\text{DP}}\) (Eq.~\ref{eq:noise_addition}) are aggregated securely via the SecAgg+ protocol \cite{bell2020secure}, ensuring confidentiality of individual contributions, even with client dropouts. The aggregated model is computed as shown in Eq.~\ref{eq:secure_aggregation} and redistributed for subsequent rounds. This iterative process continues until convergence, producing a global model ready for inference while preserving privacy.

The framework is evaluated on the BloodMNIST dataset, partitioned non-IID across \(N \in \{5, 10\}\) clients to simulate real-world data heterogeneity. Each client receives between \(0.45 \times \overline{n}\) and \(0.55 \times \overline{n}\) samples, where \(\overline{n}\) is the average sample count per client, mimicking imbalanced federated settings.

Federated training employs the FedAvg algorithm for aggregation, with each client performing \(E = 3\) local epochs per round. Privacy parameters are \((\epsilon=6.0, \delta=1.9 \times 10^{-4})\), and gradients are clipped with a norm \(C = 7\) (Eq.~\ref{eq:gradient_clipping}) to balance accuracy and privacy trade-offs. The SecAgg+ protocol uses a reconstruction threshold of four shares to ensure robustness against client dropouts. Together, these components form a secure and efficient framework for FL in medical imaging.


\section{Experiments and Results}

We evaluated our privacy-preserving FL framework on the BloodMNIST dataset under three configurations: without DP or SecAgg (DP-/SecAgg-), with only SecAgg enabled (DP-/SecAgg+), and with both techniques enabled (DP+/SecAgg+). Table~\ref{tab:results} compares our results to PriMIA \cite{kaissis2021end}, the pioneering work in this domain, and FEDMIC \cite{ren2024federated}, the current state-of-the-art.

Our experiments involved training FL models for 50 global rounds with non-IID data distributions across 10 and 20 clients to simulate real-world heterogeneity. In each configuration, local updates were secured using a fixed clipping norm (\(C=7\)) and privacy budget parameters (\(\epsilon=6.0, \delta=1.9 \times 10^{-4}\)). The SecAgg+ protocol ensured secure model aggregation, protecting individual client updates during training.

\begin{table}[h!]
  \caption{Accuracy of different configurations on the BloodMNIST dataset for varying client sizes. Results from PriMIA and FEDMIC are included for comparison.}
  \label{tab:results}
  \centering
  \begin{tabular}{llcccc}
    \toprule
    \textbf{Dataset} & \textbf{Approach} & \textbf{Client Size} & \textbf{DP-/SecAgg-} & \textbf{DP-/SecAgg+} & \textbf{DP+/SecAgg+} \\
    \midrule
    \multirow{4}{*}{BloodMNIST} 
    & \multirow{2}{*}{Ours} & 10 & 98.76 & 98.11 & 97.78 \\ 
    &  & 20 & 97.77 & 97.01 & 96.89 \\ 
    \cmidrule(r){2-6}
    & PriMIA \cite{kaissis2021end} & 10 & 90.00 & 89.00 & 85.00 \\
    \cmidrule(r){2-6}
    & FEDMIC \cite{ren2024federated} & 20 & -- & -- & 96.33 \\ 
    \bottomrule
  \end{tabular}
\end{table}

\paragraph{Conclusion}
In this work, we present a privacy-preserving federated learning framework that integrates differential privacy and secure aggregation to address privacy challenges in medical imaging. By leveraging a modified ResNet architecture tailored for DP and simulating realistic non-IID data distributions, our approach achieves superior accuracy compared to PriMIA, the pioneering work in this domain, and exceeds FEDMIC, the current state-of-the-art. Our results demonstrate the efficacy of the proposed framework, offering a robust solution for privacy-preserving medical image classification.

\bibliographystyle{plainnat} 
\bibliography{neurips_2024.bib} 
\end{document}